\pdfoutput=1

\documentclass[11pt]{article}

\usepackage[final]{acl}

\usepackage{times}
\usepackage{latexsym}

\usepackage[T1]{fontenc}

\usepackage[utf8]{inputenc}

\usepackage{microtype}

\usepackage{inconsolata}

\usepackage{graphicx}

\usepackage{amsmath}
\usepackage{booktabs}
\usepackage{multirow}
\usepackage{enumitem}
\usepackage{authblk}
\usepackage[normalem]{ulem}
\useunder{\uline}{\ul}{}
\usepackage{titlesec}
\titlespacing*{\paragraph}{0pt}{0pt}{1em}
\usepackage[most]{tcolorbox}
\usepackage{lipsum} 
\newcounter{prompt}
\renewcommand{\theprompt}{\arabic{prompt}}
\usepackage{adjustbox}

\title{Augmenting Multi-Agent Communication with State Delta Trajectory}

\author[1]{\textbf{Yichen Tang}\thanks{tangyc21@mails.tsinghua.edu.cn}}
\author[1]{\textbf{Weihang Su}}
\author[1]{\textbf{Yujia Zhou}}
\author[1]{\textbf{Yiqun Liu}}
\author[1]{\textbf{Min Zhang}}
\author[1]{\\ \textbf{Shaoping Ma}}
\author[1]{\textbf{Qingyao Ai}\thanks{Corresponding Author: aiqy@tsinghua.edu.cn}}

\affil[1]{Department of Computer Science and Technology, Tsinghua University}

\begin{document}
\maketitle
\begin{abstract}

Multi-agent techniques such as role playing or multi-turn debates have been shown to be effective in improving the performance of large language models (LLMs) in downstream tasks. 
Despite their differences in workflows, existing multi-agent systems constructed from a single base LLM mostly use natural language for agent communication.
While this is appealing for its simplicity and interpretability, it also introduces inevitable information loss as one model must down sample its continuous state vectors to discrete tokens before transferring them to the other model.
Such losses are particularly significant when the information to transfer is not simple facts, but reasoning logics or abstractive thoughts.
To tackle this problem, we propose a new communication protocol that transfers both natural language tokens and token-wise state transition trajectory from one agent to another.
Particularly, compared to the actual state value, we find that the sequence of state changes in LLMs after generating each token can better reflect the information hidden behind the inference process.
We propose a State Delta Encoding (SDE) method to represent state transition trajectories.
The experimental results show that multi-agent systems with SDE achieve SOTA performance compared to other communication protocols, particularly in tasks that involve complex reasoning.\footnote{We have open-sourced all the code and data in \url{https://github.com/LittleDinoC/StateDelta/}.}

\end{abstract}

\section{Introduction}
\label{sec:introduction}

Multi-agent systems based on Large Language Models (LLMs) have demonstrated remarkable performance in solving complex tasks~\cite{guo2024llmmultiagentsurvey, selfcollaboration, multiagentdebate}. 
While it is not surprising that combining outputs from different LLMs could improve system performance~\cite{xu2023peerreview, pre_cikm, xu-etal-2024-magic}, studies have shown that building a multi-agent system with a single base LLM can also boost the LLM’s performance~\cite{chan2023chateval,hongmetagpt, multiagentdebate}. 
These systems construct multiple agents from the same LLM, varying their profiles or access to information, which can be seen as another form of the inference scaling law~\cite{inference_scaling_laws, qian2025scalinglargelanguagemodelbased}. 
Therefore, methods for building effective multi-agent frameworks or workflows to improve LLMs in downstream tasks have been widely studied in recent literature.

Despite their differences in motivation and methodology, the majority of existing multi-agent frameworks rely on natural language tokens to build the communication protocol between agents~\cite{wu2024autogen, li2023camel, qian-etal-2024-chatdev, xieopenagents}, which may not be the optimal solution for agent communication. 
Natural language is appealing for its generalizability and interpretability, but it down samples the model’s internal states to concrete tokens before transferring information, which could lead to information loss in many cases. 
For example, in inference, an LLM may consider multiple reasoning paths, in both correct and incorrect ones could appear. 
However, only one path is ultimately sampled and presented to other agents~\cite{nlreasoningsurvey, llmrobustreasoning}, and if the sampled one is incorrect, there is no way for other agents to recover what is lost in this sampling process. 

Intuitively, when agents are built from a single base LLM (i.e., a single-LLM-based multi-agent system), information loss from natural language seems unnecessary because all agents are sharing the same semantic and parametric space created by the base LLM. 
For example, a straightforward solution to mitigate the information loss problem above is to transfer not just the final tokens, but also the token probabilities and weighted token embeddings to the other agents~\cite{cipher}.
Yet, these methods produce marginal improvements over natural language methods empirically, which indicates that simply modeling output probability distributions is not enough to convey important information hidden in the inference process of an LLM-based agent.
Thus, finding the best way to convey internal reasoning information from one agent to another has become a key research question for the studies of multi-agent communication protocols.

In this paper, we propose to augment single-LLM-based multi-agent communication directly with the model’s internal states.
Particularly, as different agents often have different initial prompts or local context in existing multi-agent frameworks, we believe that directly transferring the internal state sequence, which we refer to as the state transition trajectory, from one agent to another may not be feasible.
Instead, inspired by the idea of delta encoding~\cite{delta_encoding_for_http, delta_compression}, we propose to transfer information between agents based on both natural language tokens and the sequence of changes in the agent's internal states, which we refer to as the State Delta Encoding (SDE).
When one agent is generating output tokens, SDE records the differences between the hidden states of adjacent tokens. 
Then, when another agent is encoding these output tokens, SDE adds the trajectory of these differences (i.e., state deltas) to the corresponding tokens in order to recover the information lost in token sampling.
Our experiments on information asymmetry tasks (e.g., QA with unshared resources~\cite{dhingra2017quasar, strategyqa, complexwebquestions}) and information symmetry tasks (e.g., debates~\cite{multiagentdebate} and agent workflows~\cite{yao2023react}) show that SDE can significantly improve the performance of multi-agent systems.
The advantages of SDE are particularly strong on tasks that involve complicated logic reasoning rather than simple fact communication.
This demonstrates the potential of multi-agent communication protocols beyond natural language and could lead to multiple research directions in future studies. 

In summary, the contributions of our paper are as follows:

\begin{itemize}[leftmargin=*, itemsep=0pt, topsep=0pt]

    \vspace{1mm}

    \item We propose SDE, a novel multi-agent communication protocol that augments natural language with LLM's hidden states, bridging the gap between surface-level communication and latent reasoning. 
    
    \vspace{-0.5mm}

    \item We introduce the concept of state delta, which captures the reasoning process hidden behind output tokens and can serve as an effective medium to transfer information among single-LLM-based agents.

    \item We evaluate existing communication protocols and SDE on both information asymmetry and symmetry tasks. The results show that SDE achieves state-of-the-art performance and outperforms prior methods by up to 17.3\% in tasks that require complex reasoning.

\end{itemize}

\section{Related Work}

\subsection{LLM based Multi-Agent System}
Large Language Models have shown remarkable capabilities across a wide range of tasks~\cite{brown2020language,yang2024qwen2,fang2024scaling}, motivating researchers to explore their application in building agent systems powered by LLMs~\cite{wang2024survey,guo2024large}. 
Recent advances have shown that coordinating multiple LLM-based agents allows stronger performance in tasks such as software development~\cite{qian-etal-2024-chatdev}, world simulations~\cite{humansimulacra, li-etal-2024-econagent}, and embodied problem solving~\cite{zhangbuilding}. 

While some systems employ diverse LLMs to combine their strengths and mitigate individual biases~\cite{pre_cikm, xu2023peerreview}, many works adopt a single LLM to construct all agents, varying their behavior through different profiles or access to distinct information~\cite{qian-etal-2024-chatdev, li2023camel}. 
We refer to these as single-LLM-based multi-agent systems.
Such systems have demonstrated effectiveness through structured interactions like debates~\cite{multiagentdebate} and task-specific workflows~\cite{wu2024autogen, qian-etal-2024-chatdev}, benefiting from the scale of the inference process~\cite{inference_scaling_laws, qian2025scalinglargelanguagemodelbased}.
Our work focuses on optimizing this type of system and aims to make better use of each inference step during inter-agent communication.

\subsection{Multi-agent Communication}

Most LLM-based agent systems use natural language for communication~\cite{li2023camel, wu2024autogen, chan2023chateval}. While natural language offers flexibility, it may also introduce potential information loss.

A recent attempt to address this issue, CIPHER~\cite{cipher}, replaces natural language tokens with probability-weighted token embeddings during agent communication, showing potential in multi-agent debate settings. 
However, this approach only leverages surface-level token probability distributions from the final output layer, overlooking deeper, more informative, and more valuable hidden representations.

Another approach~\cite{ramesh2025communicatingactivationslanguagemodel} attempts to directly transfer hidden states between agents. 
In their setup, communication is restricted to a unidirectional transfer, where hidden states from a text-reading agent are passed to an output-generating agent. 
Moreover, in the multi-agent debate setting, their method requires heterogeneous models to provide distinct activations, whereas ours operates with multiple instances of the same model.

Building upon these insights, our method utilizes the dynamics of hidden states during inference and supports any inter-agent communication. 

\subsection{Latent Space Arithmetic}
Hidden states are the internal representations of LLMs that evolve token by token during generation. 
These representations encode rich semantic and contextual information, reflecting the model’s evolving understanding and planning throughout the generation process~\cite{azaria2023internal, su2024unsupervised,su-etal-2024-dragin}.
Recent studies have explored controlling the outputs of frozen LLMs by manipulating their hidden states during inference~\cite{inference_time_intervention, subramani-etal-2022-extracting}.
Several approaches have proposed extracting steering vectors to manipulate the quality~\cite{inference_time_intervention, subramani-etal-2022-extracting, rimsky-etal-2024-steering} or semantic direction~\cite{turner2024steeringlanguagemodelsactivation} of model outputs.
For example, ActAdd~\cite{turner2024steeringlanguagemodelsactivation} derives steering vectors by computing hidden state differences under prompts with or without a special keyword, and adds these vectors during inference to guide generations to a desired direction.

Inspired by these works, we also manipulate intermediate representations at inference time.
Rather than operating within a single model, we extract internal states from one agent and inject them into another.
This cross-agent state sharing aims to enhance mutual understanding and coordination in multi-agent systems.

\section{Methodology}

\begin{figure*}[h]
\centering
    \includegraphics[width=\textwidth]{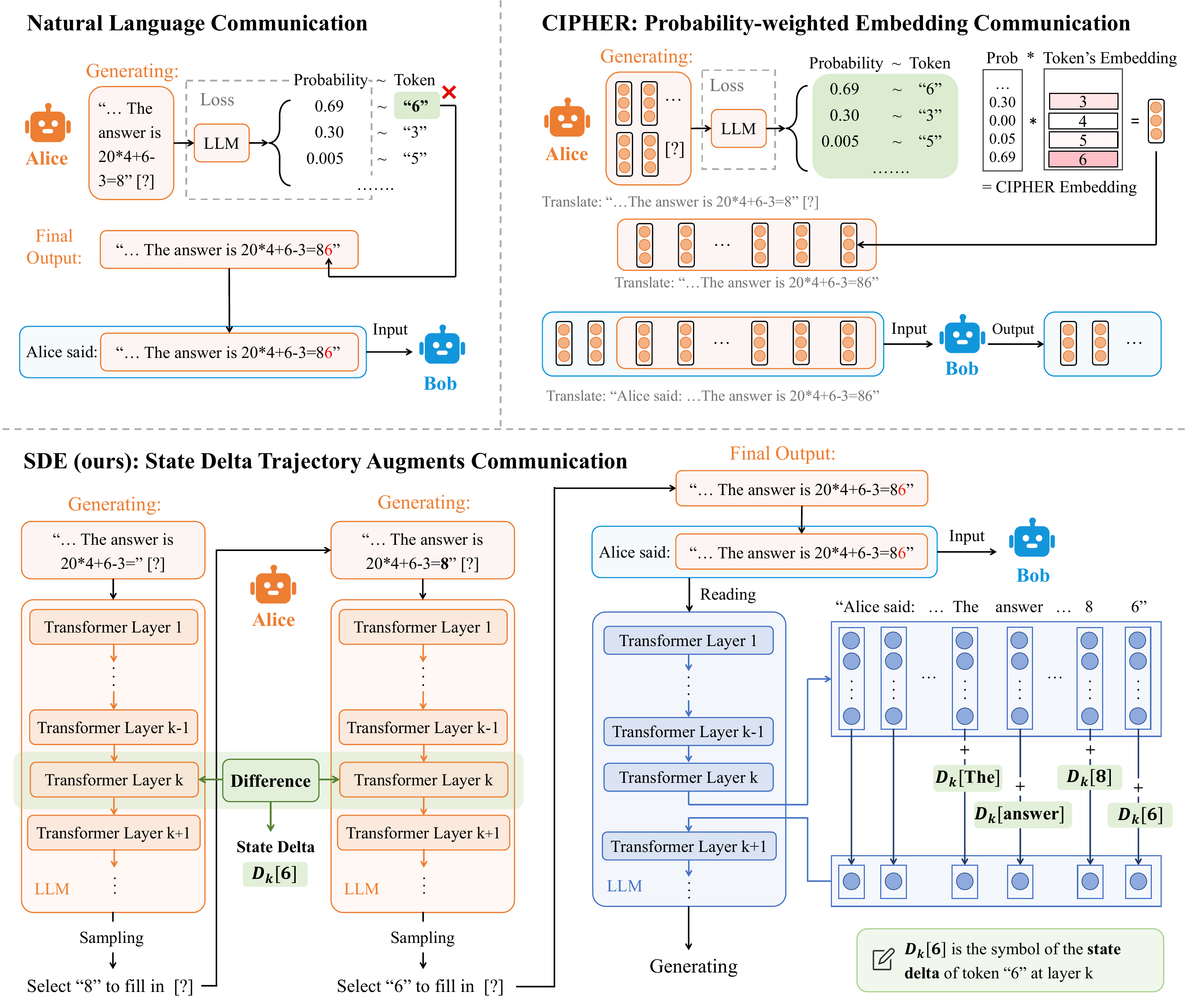}
    \caption{Comparison of different communication protocols in single-LLM-based multi-agent systems. \textbf{Top-left:} Natural language communication may introduce information loss due to sampling, thereby leading to incorrect claims being transferred. \textbf{Top-right:} CIPHER~\cite{cipher} improves by transferring probability-weighted embeddings instead of tokens, but still lacks deeper reasoning representations. \textbf{Bottom:} Our protocol (SDE) augments natural language tokens with the difference between hidden states of two adjacent tokens (state delta), bridging the gap between surface-level communication and latent reasoning.} 
    \label{pic:comparison}
\end{figure*}

We present a novel communication protocol for single-LLM-based multi-agent systems, which is constructed using a method we call \textbf{State Delta Encoding (SDE)}. 
Rather than replacing natural language, SDE augments it by transferring token-wise changes of hidden states, providing richer reasoning traces. 
This section introduces SDE as a state representation mechanism and describes how we use it to build a new communication protocol. 
The protocol with SDE is illustrated in Figure \ref{pic:comparison}.

We focus on the multi-agent systems in which all agents are constructed from the same transformer-based language model. 
Consider two agents, Alice and Bob. 
Alice receives an input and generates a response $\mathrm{output}_A$, which is a sequence of natural language tokens $t_1, t_2, t_3, \cdots, t_n$. 
In natural language communication, $\mathrm{output}_A$ is inserted directly into the input prompt of Bob. 
Formally, the prompt received by Bob, denoted as $\mathrm{prompt}_B$, takes the form $\{\mathrm{X} \  \mathrm{output}_A \mathrm{Y}\}$, where $\mathrm{X}$ and $\mathrm{Y}$ are additional contexts such as task instructions, environmental information, and responses from other agents. 
Bob then generates conditioned on $\mathrm{prompt}_B$.
However, due to sampling, the token sequence $\mathrm{output}_A$ reflects only a single reasoning path chosen by Alice, making it difficult for Bob to understand Alice's full intentions.

The inference process in causal LLMs is repeatedly performing forward propagation based on the input prompt and previously generated tokens $t_1, t_2, \cdots, t_{i-1}$ to predict the next token $t_i$. 
When Alice generates token $t_i$ in $\mathbf{output}_A$, let the hidden states $h_{A, i}^{l}$ denote the output of the $l_{th}$ transformer layer in the language model. 
Each $h_{A, i}^{l}$ is a vector representing the contextualized embedding of $t_i$, conditioned on the input prompt and previously generated tokens.
We define the state trajectory at layer $l$ during Alice's generation as the ordered sequence of hidden states:

\vspace{-2mm}

\begin{equation}
    \mathcal{H}_A^l = \{h_{A, 0}^l, h_{A, 1}^l, \cdots, h_{A, n}^l\}.
\end{equation}

\noindent Here, $h_{A, 0}^{l}$ refers to the hidden states corresponding to the last token of Alice's input prompt, serving as the initial states before generation.

As discussed in Section \ref{sec:introduction}, to prevent Bob's generation from being interfered with Alice's profile or local contexts, we avoid directly transferring the original states trajectory $\mathcal{H}_A^l$. 
Instead, inspired by the idea of delta encoding~\cite{delta_encoding_for_http, delta_compression}, we compute the differences between successive hidden states for each generated token, and define the state delta trajectory as follows:

\vspace{-3.5mm}

\begin{equation}
    \fontsize{9.5}{9}\selectfont
    \mathcal{S}_A^l = \{s_1^l, s_2^l, \cdots, s_n^l\}, \text{ where } s_i^l = h_{A, i}^{l} - h_{A, i-1}^l.    \label{eq:state_delta}
\end{equation}

\noindent Each $s_i^{l}$, referred to as a state delta, represents the internal change associated with the generation of token $t_i$. 
The state delta trajectory serves as a context-agnostic trace of the reasoning dynamics within the LLM. 
This process is called State Delta Encoding (SDE).

During communication, the state deltas serve as auxiliary information to improve Bob’s understanding of the natural language response $\mathrm{output}_A$. 
Inspired by the use of steering vectors~\cite{turner2024steeringlanguagemodelsactivation}, we treat each state delta as a steering vector and add it directly to the corresponding hidden states. 
Formally, recall that $\mathrm{prompt}_B = \{\mathrm{X} \ \mathrm{output}_A \ \mathrm{Y}\} = \{\mathrm{X}, t_1, t_2, \cdots, t_n, \mathrm{Y}\}$. 
When Bob processes $\mathrm{output}_A$ for generation, we inject the corresponding state deltas trajectory $\mathcal{S}_A^l$ into the hidden states at layer $l$ before passing them to the next layer, in order to augment each natural language token.
The hidden states $h_{B, j}^l$ of the token at position $j$ in $\mathrm{prompt}_B$ are updated as follows:

\vspace{-4.5mm}

\begin{equation}
    h_{B, j}^{l}{}' = 
    \begin{cases}
    h_{B, j}^{l} + s_{i}^{l} & \text{the position of } t_i \text{ is } j \\
    h_{B, j}^{l} & \text{otherwise} \\
    \end{cases}
    .
\end{equation}

\noindent The modified hidden states $h_B^l{}'$ are passed to the layer $l+1$ for continued inference. 
In this way, Bob not only receives the tokens, but also accesses the latent trace of Alice's internal decision-making process.
This communication protocol avoids overwriting Bob's own reasoning while guiding it to better understand Alice's generation trajectory.

\paragraph{Layer selection.} 
To minimize the impact on the model's generation capabilities, we apply SDE to only a few carefully selected layers.
The layers are chosen based on a simple preliminary experiment and remain fixed for each model throughout all subsequent tasks.
Our results show that the layers selected in this way consistently yield good results on the downstream tasks we evaluate, without requiring task-specific layer selection.
Details of our selection process are provided in Section~\ref{sec:layer_selection}.

\section{Experimental Setup}

We evaluate our approach in two settings: (1) the information asymmetry (IA) setting, where agents have access to different sets of knowledge and must collaborate to answer a question; and (2) the information symmetry (IS) setting, including multi-agent debates and agent workflows, where all agents share the same information. 
More implementation details are provided in Appendix~\ref{appendix:experiment_setup}.

\subsection{Information Asymmetry (IA) Tasks}
\label{sec:information-passing_tasks}

To simulate the cooperation process of multi-agent systems with information gaps, we propose to construct a set of information asymmetry (IA) tasks where each agent possess a unique set of information (i.e., documents) and the target task can be finished better through the collaboration of all agents. 

Specifically, we build such tasks on several factual QA benchmarks that require the system to retrieve multiple relevant documents to answer a question. 
We retrieve 6 relevant documents for each question (using a BM25-based retriever) and evenly distribute them to 2 agents as private corpora. 
To answer a target question, the agents must ask questions and respond to the questions asked by other agents based on their private corpus in order to gather the necessary information to generate the final answer. 
The agents are allowed to discuss for up to 5 rounds, and the discussion ends when either agent generates a formatted answer.

Although our passage split cannot strictly guarantee that the agents’ private information is completely disjoint, we design a single-agent baseline to demonstrate that the knowledge available to any single agent alone is insufficient to answer the questions. 
In this baseline, each agent independently performs retrieval-augmented generation using only its private passages, and we report the better of the two agents’ scores. 
The performance gap between this single-agent baseline and the multi-agent systems highlights the necessity of information exchange: only by transmitting complementary private knowledge can the agents successfully complete the task.

\paragraph{Datasets.} We evaluate our approach on three benchmarks of varied difficulty. 
(i) \textbf{Quasar-T}~\cite{dhingra2017quasar} consists of simple knowledge questions collected from various sources on the Internet. 
(ii) \textbf{ComplexWebQuestions (CWQ)}~\cite{complexwebquestions} involves multi-hop, web-based questions, which tests the model's reasoning ability over web content. 
(iii) \textbf{StrategyQA}~\cite{strategyqa} contains yes / no questions that require multi-step strategic reasoning. 
We use the first 300 questions of each dataset to build tasks.
Each question is scored by averaging over all formatted answers.
We report the average exact match (EM) scores and F1 scores in Quasar-T and ComplexWebQuestions tasks and the average accuracy in StrategyQA tasks.

\subsection{Information Symmetry (IS) Tasks}
\label{sec:thought-passing_tasks}

To evaluate how effectively agents can communicate and refine their reasoning with full information sharing, we design a set of tasks in the information symmetry (IS) setting. 
We construct two types of IS tasks: multi-agent debate and agent workflows. 
In both types, all agents have access to the same information and are required to interact by passing and refining intermediate thoughts through different structured communication frameworks.

\subsubsection{Multi-agent Debates}
\label{sec:multi-agent-debate-setup}

Inspired by~\citeauthor{multiagentdebate}, we build multi-agent debate tasks on several mathematical or logical reasoning datasets. 
At the beginning of a debate, each agent independently generates an initial answer to the same question. 
Then, in subsequent rounds, they repeatedly revise their response after observing the previous round responses of their peers. 
We simulate a 3-round debate involving 2 agents.

\paragraph{Datasets.} 
We evaluate our approach on four datasets. 
(i) \textbf{GSM8K}~\cite{cobbe2021gsm8k} contains high quality grade school math problems. 
(ii) MMLU~\cite{hendryckstest2021mmlu} is a multiple choice benchmark covering a wide range of subjects. 
we use three datasets of different categories in this benchmark: mathematical datasets \textbf{Abstract Algebra}, \textbf{College Mathematics} and logical reasoning dataset \textbf{Formal Logic}. 
We use the first 300 questions from GSM8K and the full sets of the three subsets of MMLU to build tasks.
The reported score for each question is the average accuracy of all agents' responses in the last round.

\subsubsection{Agent Workflows}
\label{sec:agent-workflow-setup}

We adapt the ReAct~\cite{yao2023react} framework to construct multi-agent workflow tasks, where agents collaborate sequentially to solve a problem by passing along thoughts and actions. 
At each step, an agent produces a thought and an action based on all previous generations, and the environment returns an observation based on the action, which becomes a part of the input for the next agent. 
Each question is solved by up to 7 agents taking turns in sequence.

\paragraph{Datasets.} 
We evaluate our approach on factual QA benchmarks and a fact verification benchmark. 
For question answering, we use two multi-hop question datasets: \textbf{HotpotQA}~\cite{yang-etal-2018-hotpotqa}, \textbf{StrategyQA}~\cite{strategyqa}. 
For fact verification, we use the \textbf{FEVER}~\cite{thorne-etal-2018-fever} dataset. 
We build tasks using the first 300 questions from each dataset. For evaluation, we report accuracy for the StrategyQA and FEVER tasks, and both EM and F1 scores for the HotpotQA task.

\subsection{Baselines}

We compare our proposed approach with the following three baselines:

\begin{itemize}[leftmargin=*, itemsep=0pt, topsep=0pt]

    \vspace{1mm}

    \item \textbf{Single}. The responses are generated by a single agent and are in natural language.
    
    \item \textbf{Natural Language (NL)}. For communication from Alice to Bob, the natural language tokens generated by Alice are inserted into Bob's input prompt.

    \item \textbf{CIPHER}~\cite{cipher}. 
    CIPHER extracts the probability distribution of each token of the corresponding forward pass, and uses this distribution to weight all tokens' embeddings, resulting in a CIPHER embedding.
    For communication from Alice to Bob, the CIPHER embedding sequences generated by Alice are inserted into Bob's input prompt in embedding form. 

\end{itemize}

NL and CIPHER use the same implementation across all tasks, while Single is implemented differently in each setting to accommodate tasks. 
We provide scenario-specific details in Appendix~\ref{appendix:experiment_setup}.

\subsection{LLM Selection and Generation Settings}

We conducted experiments on several open-source instruction-tuned LLMs. To validate the broad effectiveness of SDE, we conducted experiments with LLMs of different series on various scales, including Qwen2.5-7B-Instruct~\cite{qwen2.5}, Llama3.1-8B-Instruct~\cite{llama3.1-8b-instruct}, and Qwen2.5-14B-Instruct~\cite{qwen2.5}. 

To ensure the reproducibility of our results in IA tasks and agent workflow tasks, both NL and SDE generate responses using greedy decoding. 
Since CIPHER does not involve sampling but is affected by temperature, we set the temperature to 0 for consistency.
In multi-agent debate tasks, to promote diversity in the initial responses of different agents, we use the model's default sampling and temperature settings for generation, and all reported results are averaged over three independent runs.
More detailed settings and prompts can be found in Appendix~\ref{appendix:experiment_setup} and Appendix~\ref{appendix:prompt}.

\subsection{Layer Selection}
\label{sec:layer_selection}

We identify suitable injection layers through a simple preliminary experiment. 
Specifically, we construct an IA task using the 2WikiMultihopQA~\cite{xanh2020_2wikimultihop} dataset, following the settings described in Section~\ref{sec:information-passing_tasks}.
For each model, we evaluate SDE's performance when modifying each layer on the first 300 questions.
Considering model scales, we select 1, 2, or 3 layers for 7B, 8B, and 14B models, respectively. 
These selected layers are then used consistently across all experiments.
Notably, 2WikiMultihopQA is used only for this selection procedure and not in any main evaluation.
While the selected layers are not guaranteed to be optimal, our experiments show that they still generalize effectively to other downstream tasks we used without additional tuning.
Further analysis on the impact of different layer selections and layer counts is provided in Section~\ref{sec:ablation_on_layer_selection}. 

Detailed results and specific layer selections are reported in Appendix~\ref{appendix:layer_selection_on_2wqa}.

\begin{table*}[ht]
\caption{The experimental results in the information asymmetry tasks of SDE and other baselines on three benchmarks. The best results are in bold.}
\label{tab:crossinfo}
\centering
\vspace{-1.5mm}
{\fontsize{9.5pt}{11.4pt}\selectfont
\renewcommand{\arraystretch}{1.05}
\begin{tabular}{c|@{\hspace{4mm}}c@{\hspace{4.5mm}}|@{\hspace{0.5mm}}c@{\hspace{8mm}}c@{\hspace{5mm}}|@{\hspace{5mm}}c@{\hspace{8mm}}c@{\hspace{4mm}}|@{\hspace{4mm}}c}
\specialrule{1.0pt}{0pt}{2pt}
\rule{0pt}{11pt}
\multirow{2}{*}{\textbf{Model}} & \multirow{2}{*}{\textbf{Method}} & 
\multicolumn{2}{c}{\raisebox{0.75mm}{\hspace{-6.5mm}\textbf{Quasar-T}}} &
\multicolumn{2}{@{\hspace{-5.075mm}}|@{\hspace{5mm}}c}{\raisebox{0.75mm}{\hspace{-6mm}\textbf{CWQ}}} & 
\multicolumn{1}{@{\hspace{-4.075mm}}|@{\hspace{4mm}}c}{\raisebox{0.75mm}{\textbf{StrategyQA}}} \\

\noalign{\setlength{\arrayrulewidth}{0.5pt}}
\cline{3-7}
\noalign{\setlength{\arrayrulewidth}{1pt}}

\rule{0pt}{11pt}
&      & \hspace{4.5mm}\raisebox{-0.15mm}{\textbf{EM}}    & \raisebox{-0.15mm}{\textbf{F1}}     & \raisebox{-0.15mm}{\textbf{EM}}     & \raisebox{-0.15mm}{\textbf{F1}}     & \raisebox{-0.15mm}{\textbf{Accuracy}} \\

\specialrule{1.0pt}{2pt}{2pt}
\multirow{4}{*}{\textbf{Qwen2.5-7B-Instruct}}  & \textbf{Single} & \hspace{4.5mm}0.2367 & 0.2791 & 0.2967 & 0.3631 & 0.1700 \\
                                               & \textbf{NL}             & \hspace{4.5mm}0.3050          & 0.3748          & 0.3117          & 0.4304          & 0.4433            \\
                                               & \textbf{CIPHER}         & \hspace{4.5mm}0.2817          & 0.3567          & 0.2967          & 0.4040          & 0.3733            \\
                                               & \textbf{SDE(ours)} & \hspace{4.5mm}\textbf{0.3150} & \textbf{0.3772} & \textbf{0.3167} & \textbf{0.4444} & \textbf{0.4550}   \\

\specialrule{0.5pt}{2pt}{2pt}
\multirow{4}{*}{\textbf{Llama3.1-8B-Instruct}} & \textbf{Single} & \hspace{4.5mm}0.2333 & 0.2809 & 0.2467 & 0.3239 & 0.1500 \\
                                               & \textbf{NL}             & \hspace{4.5mm}0.2850          & 0.3496          & 0.3250          & 0.4288          & 0.4967            \\
                                               & \textbf{CIPHER}         & \hspace{4.5mm}0.2767          & 0.3488          & 0.3417          & 0.4526          & 0.5033            \\
                                               & \textbf{SDE(ours)} & \hspace{4.5mm}\textbf{0.3050} & \textbf{0.3665} & \textbf{0.3517} & \textbf{0.4640} & \textbf{0.5483}   \\
\specialrule{0.5pt}{2pt}{2pt}
\multirow{4}{*}{\textbf{Qwen2.5-14B-Instruct}} & \textbf{Single} & \hspace{4.5mm}0.3267 & 0.3845 & 0.3467 & 0.4258 & 0.4533 \\
                                               & \textbf{NL} & \hspace{4.5mm}\textbf{0.3717}   & \textbf{0.4451}  & 0.3750          & 0.4967         & 0.6733              \\
                                               & \textbf{CIPHER}         & \hspace{4.5mm}0.3517          & 0.4208          & 0.3500          & 0.4837          & 0.6433            \\
                                               & \textbf{SDE(ours)} & \hspace{4.5mm}\textbf{0.3717} & 0.4437          & \textbf{0.3817} & \textbf{0.4980} & \textbf{0.6817} \\       

\specialrule{1pt}{2pt}{0pt}
\end{tabular}}
\vspace{-1mm}
\end{table*}
\begin{table*}[ht]
\caption{The experimental results in the multi-agent debate tasks of SDE and other baselines on four benchmarks. Each reported result is the average of three independent runs. The best results are in bold.}
\centering
\small
\label{tab:debate}
\vspace{-1.5mm}
{\fontsize{9.5pt}{11.4pt}\selectfont
\renewcommand{\arraystretch}{1.05}
\begin{tabular}{c|@{\hspace{4mm}}c@{\hspace{4mm}}|@{\hspace{4mm}}c@{\hspace{4mm}}|c|c|c}
\specialrule{1.0pt}{0pt}{2pt}
\textbf{Model} & \textbf{Method} & \textbf{GSM8K} & \textbf{Abstract Algebra} & \textbf{College Math} & \textbf{Formal Logic} \\
\specialrule{1.0pt}{2pt}{2pt}
\multirow{4}{*}{\textbf{Qwen2.5-7B-Instruct}}  & \textbf{Single} & 0.8789 & 0.4767 & 0.3900 & 0.4497 \\
                                               & \textbf{NL}             & 0.9061          & 0.4583          & 0.3617          & 0.4762          \\
                                               & \textbf{CIPHER}         & 0.8933          & 0.4850          & 0.3700          & 0.4881          \\
                                               & \textbf{SDE(ours)} & \textbf{0.9178} & \textbf{0.5167} & \textbf{0.4433} & \textbf{0.5198} \\
\specialrule{0.5pt}{2pt}{2pt}
\multirow{4}{*}{\textbf{Llama3.1-8B-Instruct}} & \textbf{Single} & 0.7867 & 0.2267 & 0.2167 & 0.3571 \\
                                               & \textbf{NL}             & 0.8328          & 0.2833          & 0.2267          & 0.3889          \\
                                               & \textbf{CIPHER}         & 0.8167          & 0.2150          & 0.1950          & 0.3532          \\
                                               & \textbf{SDE(ours)} & \textbf{0.8450} & \textbf{0.3017} & \textbf{0.2417} & \textbf{0.4220} \\
\specialrule{0.5pt}{2pt}{2pt}
\multirow{4}{*}{\textbf{Qwen2.5-14B-Instruct}} & \textbf{Single} & 0.9111 & 0.5667 & 0.5067 & 0.5661 \\
                                               & \textbf{NL}             & 0.9311          & 0.7100          & 0.6350          & 0.6085          \\
                                               & \textbf{CIPHER}         & 0.9300          & 0.6500          & 0.6350          & 0.5675          \\
                                               & \textbf{SDE(ours)} & \textbf{0.9339} & \textbf{0.7533} & \textbf{0.6950} & \textbf{0.6574} \\
\specialrule{1.0pt}{2pt}{0pt}
\end{tabular}}
\vspace{-3.5mm}
\end{table*}
\begin{table}[h]
\centering
\caption{The experimental results in the agent workflow tasks of SDE and other baselines using Qwen2.5-7B-Instruct. The best results are in bold. }
\label{tab:react}
\vspace{-1.5mm}
{\fontsize{9.2pt}{11.2pt}\selectfont
\renewcommand{\arraystretch}{1.05}
\begin{tabular}{c@{\hspace{1.2mm}}|c|c@{\hspace{2.5mm}}c@{\hspace{1.9mm}}|c}
\specialrule{1.0pt}{0pt}{2pt}
\rule{0pt}{10pt}
\multirow{2}{*}{\textbf{Method}}         & 
\raisebox{0.15mm}{\hspace{0.3mm}\textbf{FEVER}}  & 
\multicolumn{2}{c}{
    \raisebox{0.15mm}{
        \textbf{HotpotQA}
    }
} & 
\multicolumn{1}{|c}{\raisebox{0.15mm}{\hspace{-0.25mm}\textbf{StrategyQA}}} \\

\noalign{\setlength{\arrayrulewidth}{0.5pt}}
\cline{2-5}
\noalign{\setlength{\arrayrulewidth}{1pt}}

\rule{0pt}{9.5pt}
  & \raisebox{-0.35mm}{\textbf{Accuracy}}     & \raisebox{-0.45mm}{\textbf{EM}}       & \raisebox{-0.45mm}{\textbf{F1}}       & \raisebox{-0.35mm}{\hspace{-1mm}\textbf{Accuracy}}         \\
\specialrule{1.0pt}{2pt}{2pt}
\textbf{Single} & 0.0067 & 0.1567 & 0.2192 & \hspace{-1mm}0.1567 \\
\textbf{NL}     & 0.2300 & 0.2100 & 0.3153 & \hspace{-1mm}0.3167 \\
\textbf{CIPHER} & 0.1800 & 0.2000 & 0.2879 & \hspace{-1mm}0.3267 \\
\textbf{SDE(ours)} & \textbf{0.2667} & \textbf{0.2267}   & \textbf{0.3196} & \hspace{-1mm}\textbf{0.3833} \\
\specialrule{1.0pt}{2pt}{0pt}
\end{tabular}}
\vspace{-6.5mm}
\end{table}

\section{Results}

\subsection{Main Experiments}
\label{sec:main_experiments}

In this section, we present the main experimental results and an analysis of our proposed SDE with other baselines in the above three tasks. 
In the following, we provide a detailed analysis of our results.

\paragraph{Overall analysis.} 
Multi-agent systems perform better than single agents directly answering in most cases. 
In particular, SDE outperforms existing communication protocols (NL and CIPHER) on almost all tasks. 
These improvements suggest that enriching communication with hidden states can indeed enhance the final collaboration performance of multi-agent systems.

Specifically, Table~\ref{tab:crossinfo} shows the results of IA tasks. 
SDE achieves a performance improvement of 0.3\% to 8.9\% compared to the best-performing baseline in most tasks, with particular notable improvements on the Llama-8B-Instruct model.
The improvements are generally more significant on multi-hop datasets CWQ and StrategyQA compared to the simple question dataset Quasar-T, indicating that SDE is more effective in tasks requiring complex, multi-step reasoning.

For the IS setting, Table~\ref{tab:debate} shows the results of multi-agent debate tasks, where SDE enhances performance ranging from 0.3\% to 13.7\% compared to the best-performing baseline.
In particular, there are significant improvements in complex mathematical and logical reasoning datasets of MMLU, where SDE consistently shows a great improvement across all evaluated models. 
Furthermore, our experiments with Qwen2.5-7B-Instruct in the agent workflow tasks (Table~\ref{tab:react}) reveal that SDE can also enhance multi-agent workflow architectures, with improvements up to 17.3\%.

\paragraph{Analysis among different tasks.} 
Results on the IA tasks demonstrate that SDE meets the fundamental requirements of communication --- accurately and effectively delivering information. 
Although SDE and NL performed similarly, the superior performance of SDE compared to CIPHER also indicates that SDE is better equipped to handle scenarios demanding higher precision in information delivery.

The more significant improvements in IS tasks indicate that SDE not only supports information delivery but also enhances agents' understanding of the reasoning processes behind the generated contents. 
This deeper comprehension boosts the overall performance of multi-agent collaboration. 

Moreover, we compare our method on StrategyQA using the same model Qwen2.5-7B-Instruct, under two different settings: information asymmetry and agent workflows. 
Our results show that the agent workflow tasks --- which require more complex reasoning --- benefit more significantly from our approach.
This also suggests that SDE is particularly effective in tasks that involve more complex reasoning processes.

\begin{figure}
\centering
\small
    \includegraphics[width=\columnwidth]{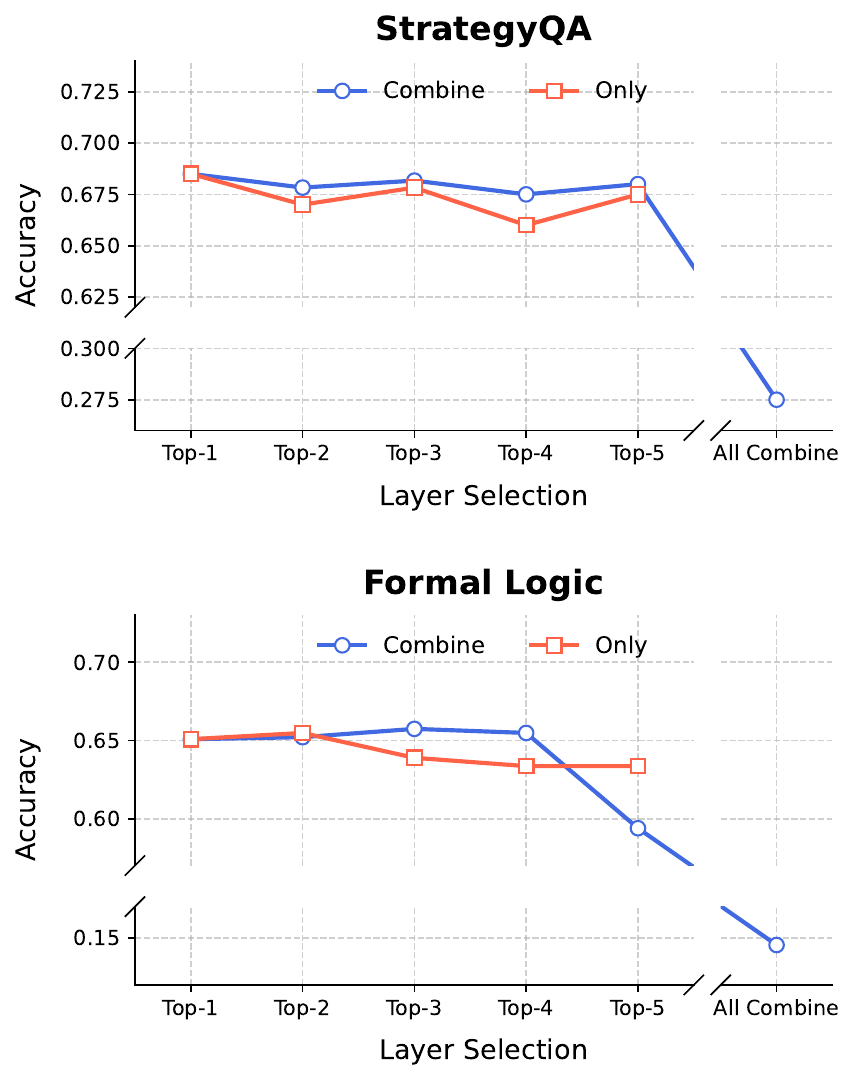}
    \vspace{-6mm}
    \caption{Ablation results for different layer selection strategies on StrategyQA (information asymmetry) and Formal Logic (multi-agent debate) tasks using Qwen2.5-14B-Instruct. We compare modifying the combined top-k layers, all layers, and only the top-k layer.} 
    \vspace{-5mm}
    \label{pic:layerselection}
\end{figure}

\subsection{Different Layer Selections}
\label{sec:ablation_on_layer_selection}

In this section, we investigate the impact of different layer selection strategies. 
Following the layer selection procedure proposed in Section~\ref{sec:layer_selection}, we compare three strategies: a combination of top-k layers, all layers, and only the top-ranking layer.

As the experiments using Qwen2.5-14B-Instruct shown in Figure \ref{pic:layerselection}, modifying the combined top-k layers (where k $\leq$ 4) results in little performance differences compared to modifying only the top-1 layer. 
At the same time, it offers greater stability than modifying a single layer.
However, modifying all layers leads to a significant performance drop, likely due to the major interference with the model's generation capabilities.
Therefore, to preserve the model's generation ability and ensure stable performance of SDE, we recommend applying the proposed layer selection procedure to the target model and modifying only a small number of top-ranking layers (e.g., 1-3).
Additional experiments on other models are provided in Appendix~\ref{appendix:layer_selection_ablation}.

\subsection{Ablation Study on State Delta}
\label{sec:ablation_on_state_delta}

To evaluate the effectiveness of the proposed state delta, we conducted an ablation study comparing the performance of the full SDE with a variant that uses the original hidden states of each token instead of their differences. 

As shown in Table~\ref{tab:state_delta_ablation}, removing state deltas consistently leads to performance drops in all settings. 
Moreover, in some cases, the performance of the variant even falls below that of using natural language alone. 
This indicates that directly augmenting with unprocessed hidden states may introduce noise, thereby impairing the agent's reasoning.

\begin{table}[t]
\centering
\small
\caption{Ablation results on the impact of state deltas in information asymmetry tasks (Quasar-T and CWQ datasets, EM scores) and multi-agent debate tasks (College Mathematics and Formal Logic datasets). "w/o delta" denotes the variant using original hidden states. 
The method with better performance is bold.}
\label{tab:state_delta_ablation}
\vspace{-1.5mm}
{\fontsize{9pt}{11pt}\selectfont
\begin{tabular}{c@{\hspace{2.7mm}}c@{\hspace{1.1mm}}c@{\hspace{2mm}}ccc}
\specialrule{1.0pt}{0pt}{1.5pt}
\textbf{} & \textbf{}   & \textbf{Quasar-T} & {\hspace{1mm}}\textbf{CWQ}    & \textbf{CM}     & \textbf{FL}     \\
\specialrule{1.0pt}{1.5pt}{1.5pt}
\multirow{3}{*}{\textbf{Q-7B}}  & \textbf{NL} & 0.3050 & 0.3117 & 0.3617 & 0.4762 \\
& \textbf{w/o delta} & 0.2950 & 0.3133 & 0.4033 & 0.4616 \\
                & \textbf{SDE} & \textbf{0.3150}  & \textbf{0.3167} & \textbf{0.4433} & \textbf{0.5198} \\
\specialrule{0.5pt}{1.5pt}{1.5pt}
\multirow{3}{*}{\textbf{L-8B}} & \textbf{NL} & 0.2850 & 0.3250 & 0.2450 & 0.3889 \\
& \textbf{w/o delta} & 0.2750 & 0.2967 & 0.2467 & 0.3942 \\
                & \textbf{SDE} & \textbf{0.3050}  & \textbf{0.3517} & \textbf{0.2967} & \textbf{0.4220} \\
\specialrule{1.0pt}{1.5pt}{0pt}
\end{tabular}}
\vspace{-2mm}
\end{table}

\subsection{Multi-agent Debate in Different Settings}

To investigate how the number of agents and rounds affects the performance in the multi-agent debate tasks, we conduct an ablation study. 
As shown in Table~\ref{tab:different_agents}, SDE consistently outperforms NL and CIPHER across different numbers of agents and rounds, suggesting that SDE is robust to variations in these configurations.

\begin{table}[t]
\centering
\small
\caption{Ablation study on the Formal Logic dataset using Qwen2.5-7B-Instruct, comparing different numbers of agents (top) and different numbers of rounds (bottom) in multi-agent debate tasks.}
\label{tab:different_agents}
\vspace{-1.5mm}
{\fontsize{9pt}{11pt}\selectfont
\begin{tabular}{ccccc}
\specialrule{1.0pt}{0pt}{1.5pt}
\textbf{Rounds} & \textbf{Agents} & \textbf{NL} & \textbf{CIPHER} & \textbf{SDE(ours)} \\
\specialrule{1.0pt}{1.5pt}{1.5pt}
3     & \textbf{2}      & 0.4762      & 0.4881          & \textbf{0.5198}         \\
3     & \textbf{3}      & 0.4489      & 0.4312          & \textbf{0.5150}         \\
3     & \textbf{4}      & 0.4530      & 0.4365          & \textbf{0.5179}         \\
3     & \textbf{5}      & 0.4947      & 0.4317          & \textbf{0.5138} \\
\specialrule{0.75pt}{1.5pt}{1.5pt}
\textbf{2}      & 2      & 0.4524      & 0.4881          & \textbf{0.5132}     \\
\textbf{3}      & 2      & 0.4762      & 0.4881          & \textbf{0.5198}     \\
\textbf{4}      & 2      & 0.4537      & 0.4881          & \textbf{0.5225}     \\
\textbf{5}      & 2      & 0.4603      & 0.4881          & \textbf{0.5212}  \\
\specialrule{1.0pt}{1.5pt}{0pt}
\end{tabular}}
\vspace{-4mm}
\end{table}

\section{Conclusion and Future Work}

In this work, we propose State Delta Encoding (SDE) and use it to build a new single-LLM-based multi-agent communication protocol.
By encoding token-wise hidden state changes, SDE captures the dynamic reasoning process during generation and reduces interference from irrelevant agent context.
The protocol with SDE augments natural language messages with token-wise state delta trajectory, enabling richer agent communication.
Experiments in both information asymmetry and symmetry tasks show that SDE outperforms existing communication protocols, especially in complex reasoning tasks. 
Our findings highlight the potential to improve communication beyond natural language and open new directions.

While our current study focuses on the application of SDE in general multi-agent settings, we believe the potential of SDE extends far beyond. 
Our future work will explore the integration of SDE into more specialized agents, such as search agents and planning agents~\cite{sun2023chatgpt,li2025search}.
In particular, for search agents, based on previous work on RAG~\cite{lewis2020retrieval,su2025parametric,su2024mitigating,su2025dynamic,dong2025decoupling,tu2025rbft} and memory~\cite{zhang2024surveymemorymechanismlarge,zhong2024memorybank}, SDE can be used to encode and track the evolution of query intent or memory state over time, enabling finer-grained control over retrieval behaviors and memory access.
We hope that extending SDE to these domains will further validate its generality and open up more possibilities for communication-aware multi-agent systems.

\section{Limitation}

While SDE shows promising improvements in multi-agent performance, it also has several limitations. 
First, SDE assumes that the hidden states of the generating agent can be easily extracted and injected into the receiving agent. 
However, this requirement may not be feasible for agents based on black-box models without internal access.
Second, incorporating hidden states increases the communication bandwidth between agents, particularly for long context communication or large models. 
Although SDE modifies only a small number of layers, this overhead may still require compression or optimization. Future work can explore selective transmission of important states or apply compression to reduce the cost of state deltas.

\section*{Acknowledgments}

This research was supported by Tsinghua University Initiative Scientific Research Program.

\bibliography{custom}

\appendix

\section{Layer Selection}
\label{appendix:layer_selection_on_2wqa}

To minimize the impact on the model’s generation capabilities, we perform layer selection to identify a small number of key transformer layers, where state deltas are captured and injected. 
We construct a preliminary experiment on an information asymmetric (IA) task using the 2WikiMultihopQA dataset~\cite{xanh2020_2wikimultihop}. 
The first 300 questions are used to evaluate each layer individually.
All settings follow those of the IA tasks, except for the dataset.
Both exact match (EM) and F1 score are used jointly as evaluation metrics. 

This procedure is applied to three models: Qwen2.5-7B-Instruct (28 layers), Llama3.1-8B-Instruct (32 layers), and Qwen2.5-14B-Instruct (48 layers).
Layers are numbered from 0.

\begin{table*}[ht]
\caption{The experimental results of the preliminary experiment constructed using the 2WikiMultihopQA dataset. Here are the Top-5 layers for SDE in each model, ranked by their combined exact match (EM) and F1 scores. Underlined layers are selected for use in all subsequent experiments.}
\label{tab:test_layer}
\centering
{\fontsize{9.5pt}{11.4pt}\selectfont
\renewcommand{\arraystretch}{1.05}
    \begin{tabular}{c@{\hspace{8mm}}c@{\hspace{10mm}}c@{\hspace{10mm}}c@{\hspace{10mm}}c@{\hspace{10mm}}c@{\hspace{10mm}}c}
    \specialrule{1.0pt}{0pt}{2pt}
      \textbf{Model} &
      \textbf{} &
      \textbf{Top 1} &
      \textbf{Top 2} &
      \textbf{Top 3} &
      \textbf{Top 4} &
      \textbf{Top 5} \\
      \specialrule{1.0pt}{2pt}{2pt}
    \multirow{3}{*}{\textbf{Qwen2.5-7B-Instruct}}  & \textbf{Layer ID} & {\ul 22} & 24       & 9      & 20     & 12     \\
                                                   & \textbf{EM}           & 0.3000   & 0.2950   & 0.3067 & 0.2900 & 0.2950 \\
                                                   & \textbf{F1}           & 0.3686   & 0.3692   & 0.3631 & 0.3703 & 0.3632 \\
      \specialrule{0.5pt}{2pt}{2pt}
    \multirow{3}{*}{\textbf{Llama3.1-8B-Instruct}} & \textbf{Layer ID} & {\ul 17} & {\ul 20} & 5      & 8      & 30     \\
                                                   & \textbf{EM}           & 0.2383   & 0.2533   & 0.2550 & 0.2417 & 0.2383 \\
                                                   & \textbf{F1}           & 0.3391   & 0.3231   & 0.3165 & 0.3168 & 0.3085 \\
      \specialrule{0.5pt}{2pt}{2pt}
    \multirow{3}{*}{\textbf{Qwen2.5-14B-Instruct}} &
      \textbf{Layer ID} &
      {\ul 33} &
      {\ul 21} &
      {\ul 23} &
      19 &
      36 \\
                                                   & \textbf{EM}           & 0.3833   & 0.3800   & 0.3817 & 0.3767 & 0.3767 \\
                                                   & \textbf{F1}           & 0.4636   & 0.4644   & 0.4585 & 0.4590 & 0.4574\\
    \specialrule{1.0pt}{2pt}{0pt}
    \end{tabular}}
\vspace{-3mm}
\end{table*}

Table~\ref{tab:test_layer} lists the top-5 layers for each model according to their combined EM and F1 scores.
Layers marked with an underline are those ultimately selected for all subsequent experiments.
The result shows that many of the top-5 layers have closely matched scores, and some even outperform the selected ones on individual metrics.
Despite variation in the exact layer rankings, we observe that the most effective layers across all models tend to be in the middle-to-late positions, for example, Layer 22 in Qwen2.5-7B-Instruct and Layer 17 in Llama3.1-8B-Instruct.
However, some earlier layers (e.g., Layers 5 and 8 in Llama3.1-8B-Instruct) also perform well, indicating potential flexibility in layer choice.

Based on this preliminary experiment, we fix the selected layers for all further experiments as follows:

\begin{itemize}[leftmargin=*, itemsep=2pt, topsep=2pt]

    \item \textbf{Qwen2.5-7B-Instruct}: Layer 22

    \item \textbf{Llama3.1-8B-Instruct}: Layers 17 and 20

    \item \textbf{Qwen2.5-14B-Instruct}: Layers 21, 23, and 33

\end{itemize}

During generation, the sender agent records state deltas from these selected layers, which are then injected into the same layers on the receiver agent's side during the forward pass. 
These layers remain fixed across all experiments to validate the generality of the selection.

It is important to note that the 2WikiMultihopQA dataset is used only in this layer selection procedure and is excluded from all evaluations.

\section{Experimental Details}
\label{appendix:experiment_setup}

All experiments were conducted using PyTorch on NVIDIA A100 GPUs with 40GB of memory. 
The specific task settings are as follows.

\subsection{Information Asymmetry (IA) Tasks}
\label{appendix:ia_tasks_details}

\paragraph{Multi-agent settings.} 
Given a factual question, two agents engage in up to five rounds of discussion to collaboratively find the answer.
We use the corpus split by DPR~\cite{karpukhin-etal-2020-dense_passage_retrieval_dpr}, including 21 million Wikipedia passages.
For each question, we retrieve the top 6 relevant passages using BM25. 
Odd-ranked passages (1st, 3rd, and 5th) are assigned to one agent, and even-ranked passages (2nd, 4th, and 6th) to another agent.
These private passages and task instructions are placed in the system prompt for each agent.

In the first round, each agent reasons based on its private knowledge and asks questions to the other agent to fill in missing information. 
In subsequent rounds, each agent receives the full responses from all agents in the previous round and is expected to respond to questions, continue reasoning, or ask new questions.
The discussion ends as soon as any agent produces a response containing an answer in the format \verb|\boxed{answer}|.
For each response in the final round, if it has such a formatted answer, we extract the answer and evaluate it.
The score of this question is calculated as the average score of all formatted answers.

Prompt~\ref{prompt:ia_ma} is used for multi-agent systems, including private passages embedded in the system prompt, the first-round response, and the second-round input that incorporates other agent's responses. 

We adopt two agents mainly to balance computational efficiency and model performance, since involving more agents substantially increases context length and may harm generation quality.
Similarly, we choose 6 passages because in comparable RAG setups a single model typically accesses three documents; in our two-agent setting, this corresponds to three private documents per agent, totaling six.

\paragraph{Single agent baseline.} 
Since each agent has different private information, we implement a single-agent answering baseline in which each agent independently performs retrieval-augmented generation based solely on its own private passages. 
We report the higher of the two agents' total scores as the baseline performance. Prompt~\ref{prompt:ia_single} is used in the single-agent baseline.

We intentionally did not evaluate a single agent with access to the union of all retrieved passages from both agents.
Our focus is on assessing whether the proposed multi-agent communication protocol can effectively transmit agent-specific private information to improve performance. 
This single-agent design aligns with that goal: the observed performance gap, where independent agents with only private information underperform compared to communicating agents, demonstrates that the agents have complementary knowledge and that successful information transmission is crucial for task completion.

\paragraph{Generation settings.} 
Each generation is limited to at most 256 tokens.
To ensure reproducibility, we use greedy decoding for Single, Natural Language (NL), and SDE methods, and set the temperature to 0 for CIPHER for fair comparison.
We note that, for CIPHER, setting temperature to 0 does not make the output equivalent to a one-hot selection. Following the official implementation\footnote{\url{https://github.com/chaudatascience/cipher\_multiagent\_debate/blob/main/models/agent.py\#L768C1-L772C1}}, when temperature is 0, the softmax is computed directly on the raw logits, which is effectively the same as setting the temperature to 1. Thus, the generated CIPHER embedding still captures the full probability distribution over the vocabulary.  

\subsection{Information Symmetry (IS) Tasks}

\subsubsection{Multi-agent Debate}

\paragraph{Multi-agent settings.} 
Given a reasoning problem, two agents engage in a three-round debate.
In the first round, each agent independently thinks through the problem and produces its initial response.
In subsequent rounds, each agent receives all other agents' responses from the previous round and is expected to revise or refine its own response based on others'.
For each question, we consider all agents’ final-round responses and calculate the task score as the proportion of correct answers among them.

Prompt~\ref{prompt:debate_ma_gsm8k} and Prompt~\ref{prompt:debate_ma_mmlu} shows an example used in the debate setting, including the first-round prompt and response, as well as the second-round prompt that incorporates the previous reply from the other agent.

\paragraph{Single agent baseline.}
We construct a single-agent baseline by providing the first-round user prompt to a single agent. 
The agent generates a single, direct response without receiving any additional inputs. This response is then used for evaluation. 
The prompt used in this single-agent setting is shown in Prompt~\ref{prompt:debate_single_gsm8k} and Prompt~\ref{prompt:debate_single_mmlu}.

\paragraph{Generation settings.}
To encourage diverse initial responses under the same first-round prompt, we use randomization during generation.
For the Single, Natural Language (NL), and SDE methods, we adopt the model's default generation settings.
In CIPHER, each agent encodes a temperature-scaled version of the model's logits into its embedding.  
Agents using the same temperature would produce identical embeddings, leading to no differences in debates.  
To address this, CIPHER assigns different temperatures to each agent to encourage varied embeddings.  
In our experiments, we follow the original implementation and set the $i$-th agent's temperature to $\frac{i}{n} \times T_\text{default}$, where $n$ is the total number of agents and $T_\text{default}$ is the model's default temperature.  
This simple scheme preserves the original CIPHER design—having one agent with a higher temperature and one with a lower temperature—without the need for dataset-specific tuning.
The default generation settings for each model are listed below:

\begin{itemize}[leftmargin=*, itemsep=2pt, topsep=2pt]
    \item Qwen2.5-7B-Instruct and Qwen2.5-14B-Instruct: \texttt{repetition\_penalty} = 1.05, \texttt{temperature} = 0.7, \texttt{top\_p} = 0.8, \texttt{top\_k} = 20
    
    \item Llama3.1-8B-Instruct: \texttt{temperature} = 0.6, \texttt{top\_p} = 0.9
\end{itemize}

\noindent To mitigate the randomness introduced by sampling, each setting is run three times and the final score is averaged between runs.
Each generation is limited to at most 512 tokens.

\subsubsection{Agent Workflow}

\paragraph{Multi-agent settings.}
In these tasks, agents sequentially generate responses in a fixed order.
Each agent receives a prompt that contains in-context examples, the current question, the complete workflow history (i.e., previous agents’ responses), and the full environmental feedback.
The agent then produces a response in a format similar to the examples, consisting of a reasoning trace (Thought) and a proposed action (Action), such as searching for documents or reporting a final answer.
The environment module validates the action and generates an observation (Observation), such as a retrieved document in response to a search action.
This observation is incorporated into the input prompt for the next agent.
We use BM25 as the retriever and Wikipedia corpus split by DPR~\cite{karpukhin-etal-2020-dense_passage_retrieval_dpr} for environment feedback in search actions.

Following the ReAct framework, each question proceeds through up to 7 iterations, with at most 7 agents contributing to the workflow.
Each agent must integrate previous reasoning and observations to refine its understanding and approach the correct answer.
The model is expected to output an answer in the format \verb|Finish[answer]|; the value of \verb|answer| is extracted for evaluation.
If no agent produces an answer in the expected format within 7 turns, the system is considered to have failed on that task.

Prompt~\ref{prompt:workflow_ma_hotpotqa} and Prompt~\ref{prompt:workflow_ma_fever} show examples of the input prompt used for multi-agent systems, including in-context examples, the first agent's reasoning and action, and the observation, all of which are provided as input to the second agent.
We adopt the examples from ReAct designed for complex reasoning (HotpotQA and StrategyQA) and fact verification (FEVER).
Due to space limitations, not all examples can be presented here. For more details, please refer to our code repository.

\paragraph{Single agent baseline.}
We construct a single-agent baseline where one agent directly answers the question by generating a chain of thought.
For the HotpotQA dataset and the StrategyQA dataset, we do not provide any retrieved documents.
For the FEVER dataset, the agent is given all possible candidate answers to choose from.
Prompt~\ref{prompt:workflow_single_hotpotqa} and Prompt~\ref{prompt:workflow_single_fever} show the prompts used for the HotpotQA / StrategyQA and FEVER datasets, respectively.

\paragraph{Generation settings.}
To ensure reproducibility, we use greedy decoding for the Single, Natural Language (NL), and SDE methods, and set the temperature to 0 for CIPHER.
For the single-agent baseline, the model’s generation is limited to 256 tokens.
For the others, we follow ReAct and limit to 100 tokens per generation.

\section{Different Layer Selections}
\label{appendix:layer_selection_ablation}

In addition to the Qwen2.5-14B-Instruct results presented in Section~\ref{sec:ablation_on_layer_selection}, we conduct further ablation studies on Qwen2.5-7B-Instruct and Llama3.1-8B-Instruct to examine different layer selection strategies.

Following the same evaluation settings as in the main experiments, we compare three strategies: (1) modifying the top-$k$ layers jointly (Combine Top-$k$), (2) modifying all layers (All Combine), and (3) modifying only the $k$-th top-ranking layer (Only Top-$k$). 
The top-$k$ layers are selected based on the preliminary experiment described in Section~\ref{sec:layer_selection} and Appendix~\ref{appendix:layer_selection_on_2wqa}. 
We evaluate these strategies on two representative tasks: an information asymmetry task based on the StrategyQA dataset and a multi-agent debate task based on the Formal Logic dataset.

\begin{figure*}[t]
\centering
    \includegraphics[width=0.85\textwidth]{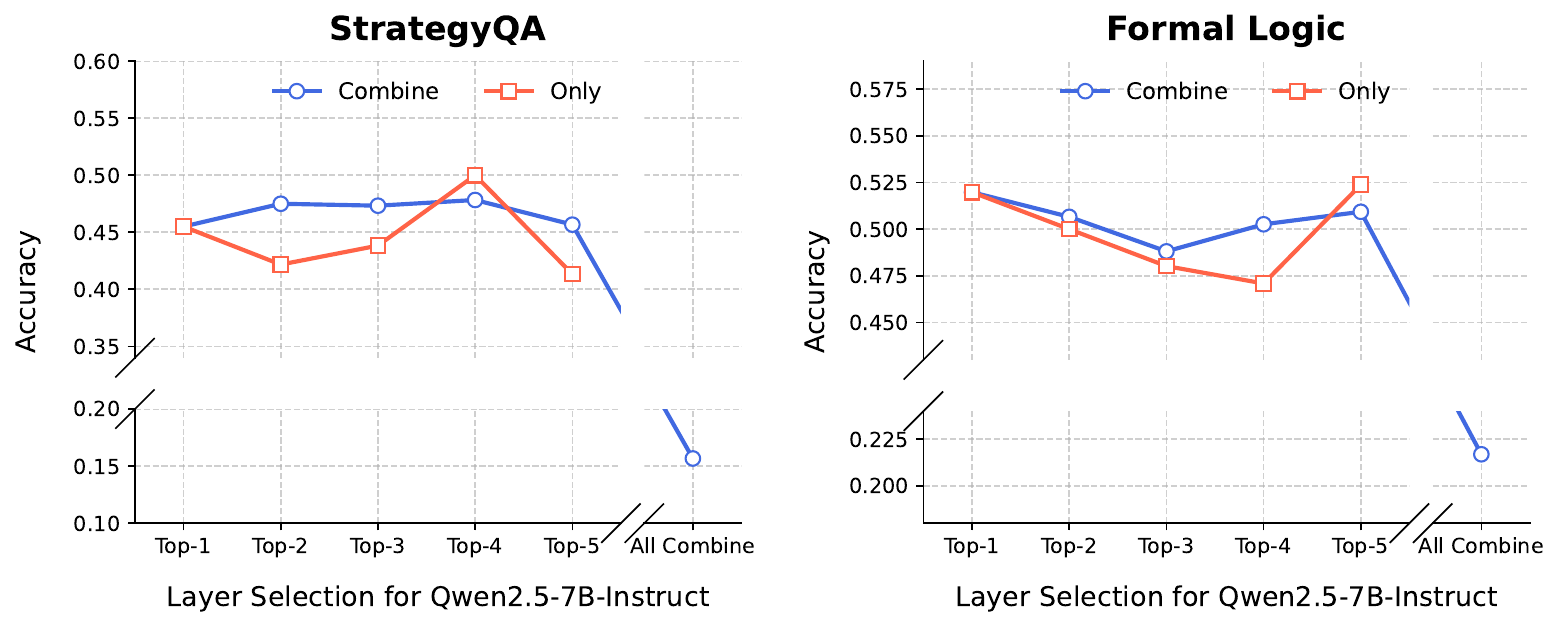}
    \caption{Ablation results for different layer selection strategies on StrategyQA (information asymmetry) and Formal Logic (multi-agent debate) tasks using Qwen2.5-7B-Instruct. We compare modifying a combination of top-k layers, all layers, and only the top-k layer.} 
    \label{pic:layer_selection_on_7b}
\end{figure*}

\begin{figure*}[t]
\centering
    \includegraphics[width=0.85\textwidth]{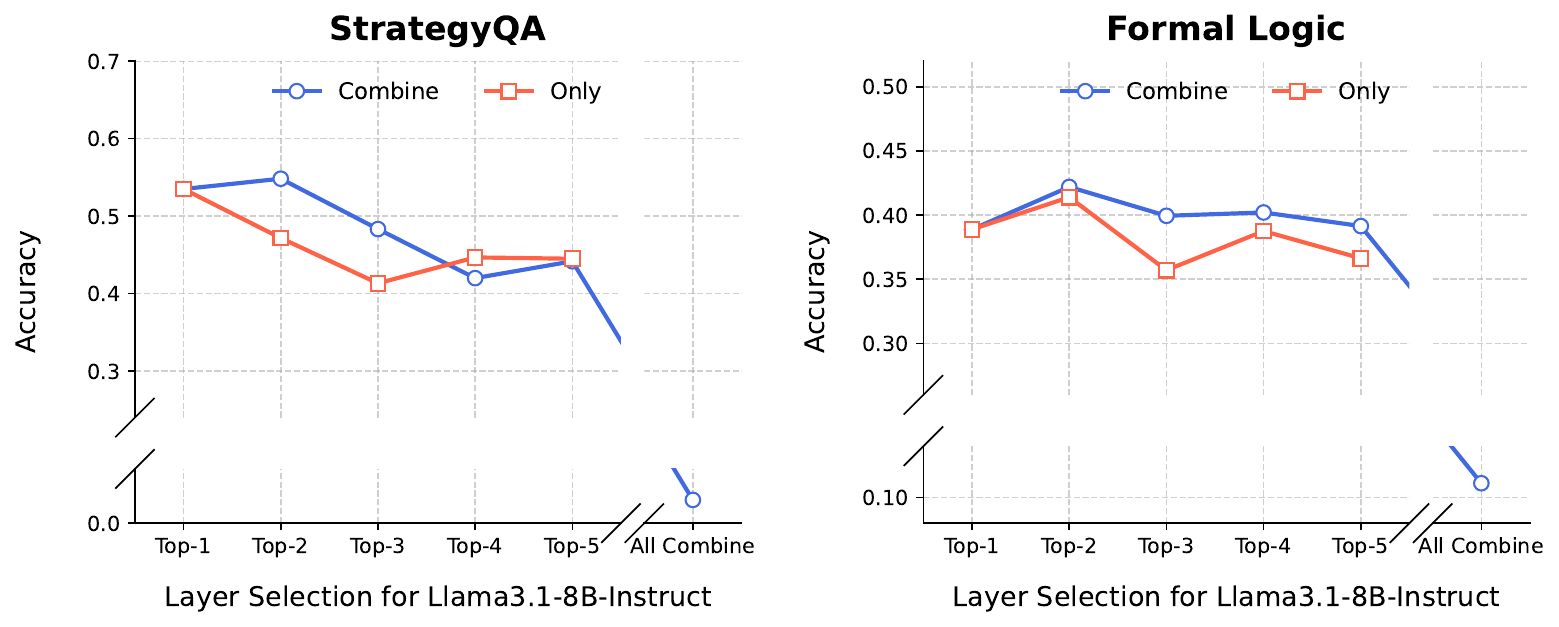}
    \caption{Ablation results for different layer selection strategies on StrategyQA (information asymmetry) and Formal Logic (multi-agent debate) tasks using Llama3.1-8B-Instruct. We compare modifying a combination of top-k layers, all layers, and only the top-k layer.} 
    \label{pic:layer_selection_on_8b}
\end{figure*}

\begin{figure*}[t]
\centering
    \includegraphics[width=0.85\textwidth]{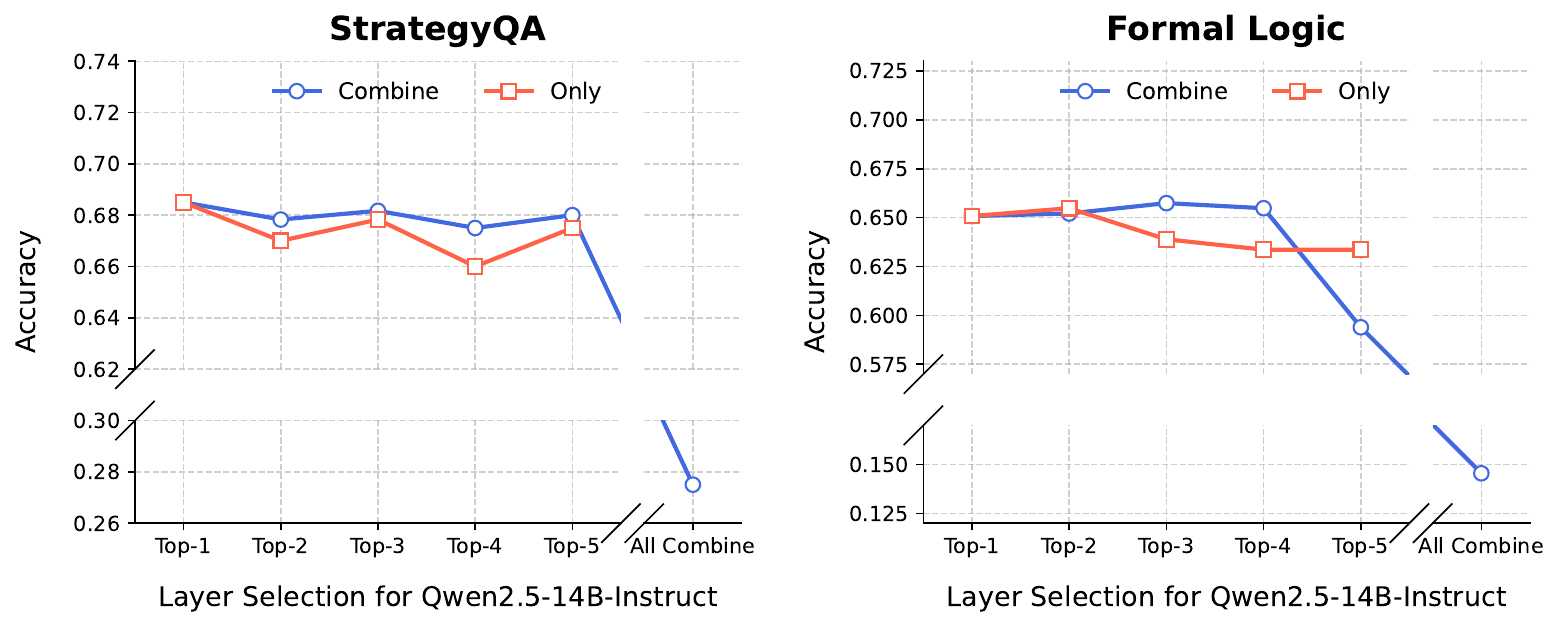}
    \caption{Ablation results for different layer selection strategies on StrategyQA (information asymmetry) and Formal Logic (multi-agent debate) tasks using Qwen2.5-14B-Instruct. We compare modifying a combination of top-k layers, all layers, and only the top-k layer.} 
    \label{pic:layer_selection_on_14b}
\end{figure*}

Figure~\ref{pic:layer_selection_on_7b}, Figure~\ref{pic:layer_selection_on_8b}, and Figure~\ref{pic:layer_selection_on_14b} show the results on Qwen2.5-7B-Instruct, Llama3.1-8B-Instruct, and Qwen2.5-14B-Instruct, respectively. 
Our key findings are as follows:

\begin{itemize}[leftmargin=*, itemsep=2pt, topsep=2pt]

\item \textbf{Single-layer modification (Only Top-$k$)} shows inconsistent performance across different layer ranks and tasks. 
For example, on Qwen2.5-7B-Instruct with the Formal Logic task, performance decreases from rank-1 to rank-4 but unexpectedly increases at rank-5. 
This suggests that single-layer modifications are sensitive to task-specific factors.

\item \textbf{Combined-layer modification (Combine Top-$k$)} yields more stable performance across different values of $k$. 
While in some isolated cases, a single-layer modification may outperform the combined version, the latter demonstrates better robustness and generality across tasks.

\item \textbf{Modifying all layers (All Combine)} consistently leads to degraded performance across all models and tasks. This is likely due to excessive disruption of the model’s internal representations, which negatively impacts its reasoning abilities.

\end{itemize}

In summary, these results further support our recommendation to apply the proposed layer selection procedure and choose a small number of combined top-ranking layers (e.g., top 1–3), avoiding the instability of single-layer selection and the performance degradation of modifying all layers.

\section{Prompts}
\label{appendix:prompt}

Here are the prompts used in our experiments.
Some complete prompts can be found in our repository.

\begin{figure*}
\centering
\refstepcounter{prompt}
\begin{tcolorbox}[colback=lightgray!20,colframe=darkgray!80,title=Prompt~\theprompt: Prompt for multi-agent systems in information asymmetry tasks, label=prompt:ia_ma]

    <system>

    You are a reasoning agent in a multi-hop problem solving task. Collaborate with other agents using these rules:

    1. **Knowledge Management**

    Your private segments:

    Document 1: \{Passage 1\}

    Document 2: \{Passage 2\}

    Document 3: \{Passage 3\}

    DO NOT verbatim share!!!

    2. **Communication Protocol**

    You can ask other agents several questions based on your needs.

    If your private segments contain information that can answer the question from other agents, you you need to give appropriate answers.

    - When asking questions:

    \ \ - First conduct reasoning based on your private segments and dialogue history

    \ \ - Identify what crucial information is missing that prevents you from progressing

    \ \ - Only ask about information you CANNOT infer from existing knowledge

    \ \ - Ask one sub-question per message

    \ \ - Never ask questions that can be answered by your own segments

    - When answering:

    \ \ - Check if the question can be answered by combining your segment with previous dialogue

    \ \ - Answer them based on your private segments

    Your communication with other agents must follow the following format:

    \textasciigrave \textasciigrave  \textasciigrave \#Q: [Your question]\textasciigrave \textasciigrave \textasciigrave 

    \textasciigrave \textasciigrave \textasciigrave \#A: [Your answer]\textasciigrave \textasciigrave \textasciigrave 

    3. **Final Output**

    When you get the final answer, response in the form \verb|\boxed{answer}| at the end of your response.

    </system>

    \vspace{3mm}

    <user>

    The multi-hop problem you need to solve collaboratively is: \{question\}

    Please communicate with other agents as required to resolve the problem. 

    </user>

    \vspace{3mm}
    
    <assistant>

    \{Agent A's response\}

    </assistant>

    \vspace{3mm}

    <user>

    Other agents responded as follows:
    
    From one agent:
    
    \{Agent B's response\}
    
    You need to answer the questions from other agents based on your private segments.
    
    The original problem is: \{question\}
    
    Please continue to think and discuss to solve this problem.
    
    When you get the final answer, response in the form \verb|\boxed{answer}| at the end of your response.

    </user>
\end{tcolorbox}
\end{figure*}
\begin{figure*}
\centering
\refstepcounter{prompt}
\begin{tcolorbox}[colback=lightgray!20,colframe=darkgray!80,title=Prompt~\theprompt: Prompt for single-agent baseline in information asymmetry tasks, label=prompt:ia_single]
    <user>
    
    Here is some relevant information:
    
    Document 1: \{Passage 1\}
    
    Document 2: \{Passage 2\}
    
    Document 3: \{Passage 2\}
    
    Please answer the following multihop question by thinking step-by-step: 
    
    \{question\}
    
    When you get the final answer, response in the form \verb|\boxed{answer}| at the end of your response.

    </user>
\end{tcolorbox}
\end{figure*}

\begin{figure*}
\centering
\refstepcounter{prompt}
\begin{tcolorbox}[colback=lightgray!20,colframe=darkgray!80,title=Prompt~\theprompt: Prompt for multi-agent systems used in multi-agent debate tasks constructed from the GSM8K dataset, label=prompt:debate_ma_gsm8k]
    <user>

    Can you solve the following math problem? \{question\}
    
    Explain your reasoning. Your final answer should be a single numerical number, in the form \verb|\boxed{answer}| at the end of your response.

    </user>

    \vspace{3mm}
    
    <assistant>

    \{Agent A's response\}

    </assistant>

    \vspace{3mm}

    <user>

    These are the solutions to the problem from other agents: 
    
    One agent solution: 
    
    \textasciigrave \textasciigrave  \textasciigrave \{Agent B's response\} \textasciigrave \textasciigrave \textasciigrave 
    
    Using the solutions from other agents as additional information, can you provide your answer to the math problem?
    
    The original math problem is \{question\}.
    
    Your final answer should be a single numerical number, in the form \verb|\boxed{answer}|, at the end of your response.
    </user>
\end{tcolorbox}
\end{figure*}
\begin{figure*}
\centering
\refstepcounter{prompt}
\begin{tcolorbox}[colback=lightgray!20,colframe=darkgray!80,title=Prompt~\theprompt: Prompt for multi-agent systems used in multi-agent debate tasks constructed from the MMLU dataset, label=prompt:debate_ma_mmlu]
    <user>
    
    Can you answer the following question as accurately as possible?
    
    \{question\}
    
    Explain your answer, putting the answer in the form (X) at the end of your response.

    </user>

    \vspace{3mm}
    
    <assistant>

    \{Agent A's response\}

    </assistant>

    \vspace{3mm}

    <user>

    These are the solutions to the problem from other agents: 

    One agent solution: 
    
    \textasciigrave \textasciigrave  \textasciigrave \{Agent B's response\} \textasciigrave \textasciigrave \textasciigrave 
    
    Using the reasoning from other agents as additional advice, can you give an updated answer? 
    Examine your solution and that other agents step by step.
    The origin question is {question}
    Put your answer in the form (X) at the end of your response.

    </user>
\end{tcolorbox}
\end{figure*}
\begin{figure*}
\centering
\refstepcounter{prompt}
\begin{tcolorbox}[colback=lightgray!20,colframe=darkgray!80,title=Prompt~\theprompt: Prompt for single-agent baseline used in multi-agent debate tasks constructed from the GSM8K dataset, label=prompt:debate_single_gsm8k]
    <user>
    
    Can you solve the following math problem? \{question\}
    
    Explain your reasoning. Your final answer should be a single numerical number, in the form \verb|\boxed{answer}|, at the end of your response.

    </user>
\end{tcolorbox}
\end{figure*}
\begin{figure*}
\centering
\refstepcounter{prompt}
\begin{tcolorbox}[colback=lightgray!20,colframe=darkgray!80,title=Prompt~\theprompt: Prompt for single-agent baseline used in multi-agent debate tasks constructed from the MMLU dataset, label=prompt:debate_single_mmlu]
    <user>
    
    Can you answer the following question as accurately as possible?
    
    \{question\}
    
    Explain your answer, putting the answer in the form (X) at the end of your response.

    </user>
\end{tcolorbox}
\end{figure*}

\begin{figure*}
\centering
\refstepcounter{prompt}
\begin{tcolorbox}[colback=lightgray!20,colframe=darkgray!80,title=Prompt~\theprompt: Prompt example for multi-agent systems used in agent workflow tasks constructed from the HotpotQA dataset and the StrategyQA dataset, label=prompt:workflow_ma_hotpotqa]
    <user>
    
    Solve a question answering task with interleaving Thought, Action, Observation steps. Thought can reason about the current situation, and Action can be three types:

    (1) Search[entity], which searches the exact entity on Wikipedia and returns the first paragraph if it exists. If not, it will return some similar entities to search.
    
    (2) Lookup[keyword], which returns the next sentence containing keyword in the current passage.
    
    (3) Finish[answer], which returns the answer and finishes the task.
    
    Here are some examples.
    
    Question: What is the elevation range for the area that the eastern sector of the Colorado orogeny extends into?
    
    Thought 1: I need to search Colorado orogeny, find the area that the eastern sector of the Colorado orogeny extends into, then find the elevation range of the area.
    
    Action 1: Search[Colorado orogeny]
    
    Observation 1: The Colorado orogeny was an episode of mountain building (an orogeny) in Colorado and surrounding areas.
    
    Thought 2: It does not mention the eastern sector. So I need to look up eastern sector.
    
    Thought 3: The eastern sector of Colorado orogeny extends into the High Plains. So I need to search High Plains and find its elevation range.
    
    Action 3: Search[High Plains]
    
    Observation 3: High Plains refers to one of two distinct land regions:
    
    Thought 4: I need to instead search High Plains (United States).
    
    Action 4: Search[High Plains (United States)]
    
    Observation 4: The High Plains are a subregion of the Great Plains. From east to west, the High Plains rise in elevation from around 1,800 to 7,000 ft (550 to 2,130 m).[3]
    
    Thought 5: High Plains rise in elevation from around 1,800 to 7,000 ft, so the answer is 1,800 to 7,000 ft.
    
    Action 5: Finish[1,800 to 7,000 ft]

    ......(five examples omitted)......
    
    Question: Were Scott Derrickson and Ed Wood of the same nationality?

    Thought 1: I need to determine the nationalities of Scott Derrickson and Ed Wood to see if they are the same. I'll start by searching for Scott Derrickson.

    Action 1: Search[Scott Derrickson]

    Observation 1: Scott Derrickson is an American director, screenwriter and producer. He lives in Los Angeles, California. Derrickson is best known for directing numerous horror films, such as \"The Exorcism of Emily Rose\" (2005), \"Sinister\" (2012), and \"Deliver Us From Evil\" (2014), as well as the Marvel Cinematic Universe superhero film \"Doctor Strange\" (2016). He was born on July 16, 1966 in Denver, Colorado, United States. He is married to actress and producer Jennifer Derrickson, with whom he has two children.

    </user>

    \vspace{3mm}

    <assistant>

    Thought 2: 
\end{tcolorbox}
\end{figure*}
\begin{figure*}
\centering
\refstepcounter{prompt}
\begin{tcolorbox}[colback=lightgray!20,colframe=darkgray!80,title=Prompt~\theprompt: Prompt example for multi-agent systems used in agent workflow tasks constructed from the FEVER dataset, label=prompt:workflow_ma_fever]
    <user>
    
    Determine if there is Observation that SUPPORTS or REFUTES a Claim, or if there is NOT ENOUGH INFORMATION.
    
    Claim: Nikolaj Coster-Waldau worked with the Fox Broadcasting Company.
    
    Action 1: Search[Nikolaj Coster-Waldau]
    
    Observation 1: Nikolaj William Coster-Waldau (born 27 July 1970) is a Danish actor and producer. He graduated from the Danish National School of Performing Arts in Copenhagen in 1993,[1] and had his breakthrough role in Denmark with the film Nightwatch (1994). He played Jaime Lannister in the HBO fantasy drama series Game of Thrones, for which he received two Primetime Emmy Award nominations for Outstanding Supporting Actor in a Drama Series.. Coster-Waldau has appeared in numerous films in his native Denmark and Scandinavia, including Headhunters (2011) and A Thousand Times Good Night (2013). In the U.S, his debut film role was in the war film Black Hawk Down (2001), playing Medal of Honor recipient Gary Gordon.[2] He then played a detective in the short-lived Fox television series New Amsterdam (2008), and appeared in the 2009 Fox television film Virtuality, originally intended as a pilot.
    
    Action 2: Finish[SUPPORTS]

    Claim: Stranger Things is set in Bloomington, Indiana.
    
    Action 1: Search[Stranger Things]
    
    Observation 1: Stranger Things is an American science fiction horror drama television series created by the Duffer Brothers. Set in the 1980s, primarily in the fictional town of Hawkins, Indiana, the series centers on a number of mysteries and supernatural events occurring around the town and their impact on an ensemble of child and adult characters.
    
    Action 2: Finish[REFUTES]

    Claim: Beautiful reached number two on the Billboard Hot 100 in 2003.?
    
    Action 1: Search[Beautiful]
    
    Observation 1: Could not find [Beautiful]. Similar: ['Beautiful', 'Beautiful, Beautiful', 'A Beautiful Mind (film)', 'Beautiful (Christina Aguilera song)', 'Life Is Beautiful'].
    
    Action 2: Search[Beautiful (Christina Aguilera song)]
    
    Observation 2: "Beautiful" is a song recorded by American singer Christina Aguilera for her fourth studio album, Stripped (2002).
    
    Action 3: Lookup[Billboard Hot 100]
    
    Observation 3: (Result 1 / 3) The song peaked at number two on the Billboard Hot 100 in the United States, where it was certified Gold for 500,000 units shipped.
    
    Action 4: Finish[NOT ENOUGH INFO]

    
    Claim: There is a convicted statutory rapist called Chinatown's writer.

    Thought 1: To determine if this claim is supported, refuted, or if there is not enough information, I need to search for information about a convicted statutory rapist named "Chinatown's writer."

    Action 1: Search["Chinatown's writer" convicted statutory rapist]

    Observation 1: bed linens. Those investigating serial rapes often identify the rapist with a 'nickname' before an arrest is made by characterizing the tactics or patterns of the rapes. Serial rapists are more likely to be convicted than a rapist who is known by the victim. Unlike those convicted for a single case of rape, serial rapists often go unrecognized due to the slow process of analyzing the backlog of rape kits. It may take many years for a past rape to be identified as being committed by one person.

    </user>

    \vspace{3mm}

    <assistant>

    Thought 2: 
\end{tcolorbox}
\end{figure*}
\begin{figure*}[t]
\centering
\refstepcounter{prompt}
\begin{tcolorbox}[colback=lightgray!20,colframe=darkgray!80,title=Prompt~\theprompt: Prompt for single agent baseline used in agent workflow tasks constructed from the HotpotQA dataset and the StrategyQA dataset, label=prompt:workflow_single_hotpotqa]
    <user>
    
    Please answer the following multihop question by thinking step-by-step: 
    
    \{question\}
    
    When you get the final answer, response in the form \verb|\boxed{answer}| at the end of your response.

    </user>
\end{tcolorbox}
\end{figure*}
\begin{figure*}[t]
\centering
\refstepcounter{prompt}
\begin{tcolorbox}[colback=lightgray!20,colframe=darkgray!80,title=Prompt~\theprompt: Prompt for single agent baseline used in agent workflow tasks constructed from the FEVER dataset, label=prompt:workflow_single_fever]
    <user>
    
    Please answer the following multihop question by thinking step-by-step: 
    
    There is a convicted statutory rapist called Chinatown's writer.
    
    When you get the final answer, response in the form \verb|\boxed{answer}| at the end of your response.
    
    All final answers can only be one of "NOT ENOUGH INFO", "SUPPORTS", "REFUTES".

    </user>
\end{tcolorbox}
\end{figure*}

\end{document}